\documentclass[10pt,twocolumn]{article}

\usepackage[letterpaper,margin=0.65in]{geometry}
\usepackage[T1]{fontenc}
\usepackage{lmodern}
\usepackage{microtype}
\usepackage{amsmath}
\usepackage{amssymb}
\usepackage{graphicx}
\usepackage{booktabs}
\usepackage{tabularx}
\usepackage{natbib}
\usepackage[hidelinks]{hyperref}
\graphicspath{{figures/}}
\title{Do Time-Series Foundation Models Pay Off for Industrial Monitoring?\\
A Cost-Aware Empirical Study}
\author{Guan-Hua Wen \quad Kuan-Yu Chen\\
Department of Computer Science and Information Engineering\\
National Taiwan University of Science and Technology\\
Taipei, Taiwan\\
\texttt{B11215005@mail.ntust.edu.tw, kychen@mail.ntust.edu.tw}}
\date{}

\begin{document}
\maketitle

\begin{abstract}
Industrial monitoring models must detect operationally relevant deviations while satisfying target-specific data, calibration, and resource constraints.
Time-series foundation models (TSFMs) promise reusable representations and zero-shot forecasts, yet evidence for their deployment value remains mixed when task definitions are heterogeneous and lightweight baselines are competitive.
This work presents a protocol-aware empirical assessment across three settings: a C-MAPSS degradation-risk proxy, normal-only training for anomalous-sound detection on MIMII, and BDG2 forecasting-residual diagnostics with synthetic target perturbations.
We assess classical one-class methods, compact neural autoencoders, residual forecasters, MOMENT-small, Chronos-T5, and TimesFM~2.5 in terms of anomaly-ranking performance, risk-horizon sensitivity, residual forecasting and perturbation sensitivity, and local implementation cost.
Across 100 C-MAPSS engines evaluated out of fold, TCN-AE reaches fold-weighted AUROC/AUPRC 0.9570/0.8960, compared with 0.7310/0.3080 for MOMENT reconstruction; paired engine-cluster bootstrap confidence intervals exclude zero for both differences.
Across five matched MIMII pump evaluations, OCSVM also exceeds MOMENT reconstruction in AUROC and AUPRC.
On a fixed 12-meter BDG2 panel, TimesFM~2.5 has the lowest aligned forecast error and the highest synthetic AUROC point estimate, although synthetic AUPRC is similar across TSFM and fitted residual models.
Same-device measurements show that MOMENT incurs higher latency, peak allocated VRAM, and serialized state-dictionary size than TCN-AE.
Under the evaluated frozen and zero-shot settings, TSFMs are task-dependent deployment options rather than default replacements for fitted lightweight models.
\end{abstract}

\section{Introduction}
\label{sec:introduction}

Industrial monitoring supports decisions that range from maintenance scheduling to fault investigation and energy-system operation.
Useful models must match target labels and evaluation units while meeting local resource constraints.
TSFMs promise representations or forecasts that transfer with little target-specific adaptation \citep{goswami2024moment,ansari2024chronos,das2024timesfm}, but their deployment value is uncertain when target-normal data and competitive lightweight models are available.

The target tasks do not share a native anomaly definition.
C-MAPSS provides run-to-failure trajectories, MIMII supports normal-only acoustic anomaly detection, and the BDG2 data used here lack real-fault labels \citep{saxena2008cmapss,purohit2019mimii,koizumi2020dcase,miller2020bdg2}.
We therefore compare models only within protocol, using domain-specific evaluation units, metrics, and uncertainty procedures rather than a cross-domain leaderboard.
The study asks four deployment questions:
\begin{enumerate}
    \setlength{\itemsep}{2pt}
    \setlength{\parsep}{0pt}
    \setlength{\topsep}{2pt}
    \item Under matched engine- or file-level protocols, do frozen TSFM modes improve anomaly ranking over fitted lightweight models?
    \item Does the C-MAPSS model ordering persist when the degradation-risk horizon changes?
    \item Do zero-shot forecasting TSFMs improve aligned BDG2 forecasting residuals and sensitivity to controlled perturbations?
    \item What latency, memory, and model-size costs accompany any predictive gain under a standardized local workload?
\end{enumerate}

Across classical estimators, compact neural models, and three TSFM families, we contribute matched C-MAPSS and MIMII comparisons, a prespecified multi-meter BDG2 diagnostic, grouped uncertainty analyses, and standardized local cost metadata.
This study provides an empirical deployment comparison; it does not propose a new detector, claim state-of-the-art performance, or establish a universal ranking of model families.

\section{Related Work}
\label{sec:related-work}

\paragraph{Time-series anomaly detection.}
Recent surveys organize deep approaches around reconstruction, forecasting, density estimation, contrastive learning, and graph- or attention-based architectures \citep{darban2024deep}.

\paragraph{Industrial transfer and domain shift.}
Changes in operating regime, machine identity, or sensing conditions can degrade a detector fitted to a source distribution.
Deep transfer-learning surveys emphasize the potential to reduce retraining and labeling requirements, while warning that source--target mismatch must be handled explicitly \citep{yan2023transfer}.

\paragraph{Acoustic machine anomaly detection.}
MIMII was introduced to support machine-sound anomaly detection under realistic factory noise, with normal and anomalous recordings from valves, pumps, fans, and slide rails \citep{purohit2019mimii}.
DCASE 2020 Task 2 formalized unsupervised anomalous sound detection in which only normal sounds are available for training and unknown abnormal sounds must be detected at test time \citep{koizumi2020dcase}.
Our experiments adopt the normal-only training premise of DCASE Task 2 but do not reproduce its complete machine-ID-specific scoring protocol.

\paragraph{Building energy residual monitoring.}
BDG2 contains large-scale building meter data associated with the ASHRAE Great Energy Predictor III context \citep{miller2020bdg2}.
Because the BDG2 data used here do not provide equipment-fault labels, we report forecasting errors and separately use injected perturbations to probe residual sensitivity.

\paragraph{Time-series foundation models.}
MOMENT introduced an open family of pretrained models for forecasting, classification, anomaly detection, and other time-series tasks \citep{goswami2024moment}.
Chronos adapts language-model architectures to probabilistic forecasting by scaling and quantizing time-series values into tokens \citep{ansari2024chronos}.
TimesFM uses a decoder-only architecture for zero-shot forecasting \citep{das2024timesfm}.
Recent work such as ChronosAD explores foundation-model representations specifically for anomaly detection \citep{khan2026chronosad}.
An empirical assessment in power-system forecasting evaluates TSFMs across zero-shot accuracy, fine-tuning efficiency, horizon sensitivity, covariate handling, and unseen-site generalization \citep{zater2026empirical}.
We instead assess frozen or zero-shot deployment modes across matched industrial monitoring protocols and local implementation costs.

\section{Methods}
\label{sec:benchmark}

\subsection{Task Framing}

The three domains do not share a native anomaly definition.
Treating them as entries in one leaderboard would therefore be misleading.
Instead, we use a common interface,
\[
x \longmapsto s(x) \in \mathbb{R},
\]
mapping each input sequence or window to a scalar monitoring score while preserving domain-specific interpretation.
Table~\ref{tab:protocols} summarizes the resulting protocols.

For C-MAPSS, multivariate trajectories are divided into sliding windows.
For engine $i$ with terminal cycle $T_i$, a window ending at cycle $t$ receives
\[
\operatorname{RUL}_{i,t}=T_i-t,
\qquad
y_{i,t}^{(h)}=\mathbf{1}\{\operatorname{RUL}_{i,t}\le h\}.
\]
The main analysis uses $h=30$ cycles and evaluates $h\in\{15,20,30,40\}$ for sensitivity.
This is a degradation-risk proxy, not a native anomaly label.
For MIMII, models are calibrated on normal training audio and evaluated on test-normal and test-anomaly files.
For BDG2, a forecasting model predicts electricity readings and absolute residual magnitude provides the monitoring score.
Synthetic perturbations create pseudo labels only for controlled sensitivity analysis.

\begin{table*}[t]
\centering
\caption{Protocol summary. C-MAPSS uses degradation-risk proxy labels, MIMII uses native file-level test labels, and BDG2 uses synthetic perturbation diagnostics. Results are interpreted within each protocol and are not compared across datasets.}
\label{tab:protocols}
\small
\setlength{\tabcolsep}{5.5pt}
\begin{tabularx}{\textwidth}{@{}l l X X >{\raggedright\arraybackslash}p{0.14\textwidth}@{}}
\toprule
Dataset & Domain & Task framing & Training/evaluation protocol & Primary metrics \\
\midrule
C-MAPSS & Turbofan sensors & Degradation trajectories converted to a high-risk-window proxy & Five engine-disjoint folds; 512 proxy-normal fitting windows per fold; 100 engines evaluated out of fold & AUROC, AUPRC, F1, precision, recall \\
MIMII & Machine acoustics & Labeled anomalous sound detection & Five paired file manifests; 160 train-normal and 320 balanced test files per seed; four machine IDs & AUROC, AUPRC, F1, precision, recall \\
BDG2 & Building energy & Forecast residual monitoring and synthetic sensitivity & Fixed 12-meter panel; aligned held-out timestamps; five injection seeds and ten perturbation types & MAE, RMSE, sMAPE; pseudo AUROC, AUPRC, F1 \\
\bottomrule
\end{tabularx}
\end{table*}

\subsection{Datasets and Preprocessing}

\paragraph{C-MAPSS.}
C-MAPSS contains simulated turbofan degradation trajectories generated with NASA's Commercial Modular Aero-Propulsion System Simulation tool \citep{saxena2008cmapss}.
Each row represents one operating cycle with three operating settings and 21 sensor measurements \citep{saxena2008cmapss}.
We use the 21 sensor channels, exclude the three operational settings from the model input, and construct length-30 windows with stride 1.
A seed-controlled five-fold manifest assigns each of the 100 FD001 engines to exactly one held-out fold; every fold contains 80 fitting and 20 test engines, with zero engine or sample-ID overlap.
Within each fold, 512 proxy-normal windows are selected by a seed-controlled sample-ID rule shared by all models.
Per-sensor standardization is fitted only on these selected windows, and every window from the held-out engines is evaluated.

\paragraph{MIMII.}
We use the pump archive from the MIMII/DCASE machine-sound data \citep{purohit2019mimii,koizumi2020dcase}.
Audio is preprocessed once into a shared cache of fixed-size $128\times32$ log-mel tensors using 32 mel bins, FFT size 1024, and hop length 512.
For each of five seeds, a manifest selects 160 pump train-normal files and 320 balanced test files across machine IDs 00, 02, 04, and 06.
Normalization is fitted on the train-normal tensors only.
MOMENT consumes the normalized tensors directly; classical models consume a deterministic flattening of the same tensors.
All anomaly scores and metrics remain file-level.

\paragraph{BDG2.}
We use cleaned hourly electricity meter series from BDG2 \citep{miller2020bdg2}.
We prespecify a 12-meter electricity panel selected with seed 42 from eligible meters having at least 12,000 observations, at least 90\% coverage, and nonzero variance.
The panel spans nine sites and five primary-use categories.
Lightweight regressors use calendar features and one- and 24-hour lags.
Chronos and TimesFM receive a 48-hour historical context and make zero-shot one-step forecasts.
Forecasts are aligned to the common held-out timestamps available from every model.

\subsection{Anomaly Scores and Thresholds}

All evaluated monitoring methods return scores for which larger values indicate greater deviation under the corresponding protocol.
Classical one-class models use the negated estimator score, reconstruction models use mean squared reconstruction error, and forecasting models use absolute residuals.
Unless otherwise noted, the operating threshold is the 95th percentile of training scores:
\[
\tau = Q_{0.95}\left(\{s(x_i):x_i\in\mathcal{D}_{\mathrm{train}}\}\right).
\]
This avoids selecting decision thresholds using test labels.

\subsection{Model Families}

\paragraph{Classical one-class baselines.}
We evaluate Isolation Forest \citep{liu2008isolation}, One-Class SVM \citep{scholkopf2001ocsvm}, Local Outlier Factor in novelty mode \citep{breunig2000lof}, and PCA reconstruction \citep{jackson1979pca}.
Windowed sensor inputs are flattened before model fitting.

\paragraph{Lightweight deep baselines.}
C-MAPSS uses compact LSTM \citep{hochreiter1997lstm} and TCN-style \citep{bai2018tcn} autoencoders.
Both are trained on scaled normal windows, and per-window reconstruction error is used as the anomaly score.

\paragraph{Energy residual baselines.}
We compare lag-24 naive forecasts, rolling medians, Ridge regression \citep{hoerl1970ridge}, and LightGBM \citep{ke2017lightgbm}.
These models are evaluated using forecasting residual metrics and, separately, synthetic pseudo-anomaly diagnostics.

\paragraph{Foundation models.}
MOMENT-small \citep{goswami2024moment} is evaluated in embedding-plus-Isolation-Forest and reconstruction modes for C-MAPSS; reconstruction is also evaluated on fixed-length MIMII log-mel windows.
Chronos-T5 tiny \citep{ansari2024chronos} and TimesFM~2.5 200M \citep{das2024timesfm,google2025timesfm25} are evaluated as zero-shot one-step BDG2 forecasters.
The selected TSFMs operate in different task modes and domains; the study compares deployment cases rather than estimating a domain-independent ranking of model families.
The selection is not exhaustive and does not cover fine-tuning or anomaly-specific adapters.

\subsection{Cost and Reproducibility Metadata}

Each formal run stores protocol and train-selection checksums together with Parquet score artifacts, allowing summary metrics to be recomputed and paired by engine or file.
The standardized C-MAPSS cost audit uses one RTX 4090, float32 inference, identical scaled fold-0 windows, batch sizes 1 and 16, 10 warm-up runs, and 30 measured repetitions.
It reports same-process warm-call p50/p95 per-batch model-call latency, peak allocated/reserved VRAM, and serialized state-dictionary size.
Preprocessing and model loading are excluded and reported as such; CPU Isolation Forest is retained as a deployment configuration but is not used for same-device algorithmic dominance.

Dataset archives are tracked by path, size, and SHA256 checksum.
Experiments use YAML configurations, and paper artifacts are regenerated from run-level metrics.
The main Python~3.11 environment is locked with \texttt{uv.lock}; TimesFM~2.5 uses a separately specified environment because of incompatible NumPy requirements.

\subsection{Evaluation Units}

C-MAPSS uses engines as the split and bootstrap unit; overlapping windows from an engine remain in the same fold.
MIMII uses audio files as the evaluation unit, and all compared models consume the same file manifests and cached tensors within a seed.
BDG2 uses meters as the primary aggregation unit and injection realizations nested within meters.
No result is interpreted as raw transfer across sensor, acoustic, and energy inputs.

\subsection{Uncertainty}

For C-MAPSS, point estimates are engine-count-weighted means of fold metrics; paired differences use 10,000 bootstrap resamples of held-out engines within each fold, avoiding direct ranking of raw scores across fold-specific scales.
For MIMII, we report means and $t$-based 95\% confidence-interval half-widths over five paired file manifests; paired file-bootstrap diagnostics resample within label-by-machine-ID strata.
BDG2 first averages injection seeds within a meter and then reports meter-macro means and dispersion.

\subsection{TSFM Evaluation and Synthetic Diagnostics}

Each TSFM uses the matched domain inputs and evaluation units defined above.
Chronos-T5 and TimesFM forecast each held-out BDG2 target from a fixed context.

Five injection seeds place nonoverlapping six-point events at nominal point rate 0.01 across spike, drop, flatline, gradual-drift, level-shift, intermittent-missingness, stuck-at, delayed-response, seasonal-distortion, and contextual-anomaly types while forecasts and contexts remain fixed.
Type-specific AUROC, AUPRC, F1, precision, and recall are reported alongside aggregate diagnostics.
Directly observable missingness is a separate pipeline check excluded from aggregate metrics.
These pseudo labels diagnose residual sensitivity, not equipment faults, occupancy events, commissioning issues, or meter failures.

\section{Results}
\label{sec:results}

\subsection{Target-Domain Anomaly Ranking}

\begin{figure*}[t]
\centering
\includegraphics[width=0.88\textwidth]{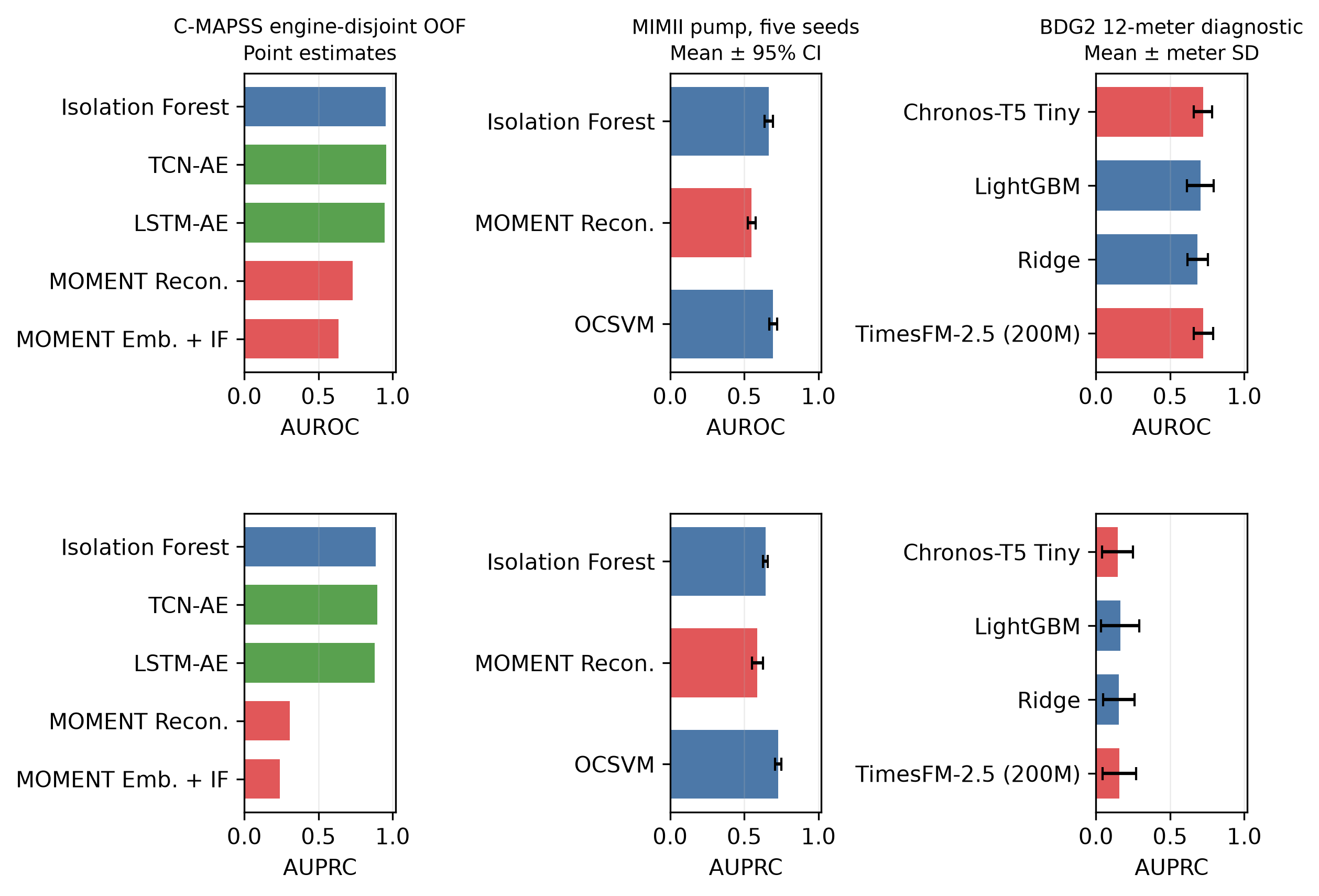}
\caption{Protocol-specific AUROC and AUPRC; values should not be compared across datasets. Blue denotes classical or fitted baselines, green compact neural models, and red TSFMs. C-MAPSS reports engine-count-weighted means from engine-disjoint folds. MIMII reports five-seed means with $t$-based 95\% confidence-interval half-widths. BDG2 reports meter-macro means with across-meter standard deviations after averaging injection seeds within each meter; these values measure synthetic perturbation sensitivity.}
\label{fig:protocol-performance}
\end{figure*}

Figure~\ref{fig:protocol-performance} and Table~\ref{tab:representative-results} summarize the matched-protocol results.
Across 100 held-out C-MAPSS engines and 17,731 out-of-fold windows, TCN-AE obtains fold-weighted AUROC 0.9570, AUPRC 0.8960, and F1 0.7959.
Isolation Forest is close at 0.9521/0.8866/0.7847, whereas the better-performing MOMENT configuration, reconstruction, reaches only 0.7310 AUROC and 0.3080 AUPRC.

The fold-stratified engine-cluster paired bootstrap supports the C-MAPSS gap.
The paired differences, defined as MOMENT minus TCN-AE, are $-0.2260$ for reconstruction AUROC and $-0.5880$ for reconstruction AUPRC; the corresponding embedding differences are $-0.3199$ and $-0.6570$.
The respective 95\% confidence intervals from 10,000 within-fold engine resamples are $[-0.2557,-0.1938]$, $[-0.6226,-0.5260]$, $[-0.3458,-0.2945]$, and $[-0.6886,-0.6235]$.

Under the paired MIMII protocol, OCSVM has the strongest five-seed AUROC/AUPRC means, $0.6944\pm0.0272$ and $0.7277\pm0.0209$, compared with $0.5501\pm0.0271$ and $0.5890\pm0.0367$ for MOMENT reconstruction.
Within every seed, the label-by-machine-ID-stratified paired file-bootstrap 95\% confidence intervals for MOMENT minus OCSVM lie below zero for both AUROC and AUPRC.
All methods share files, upstream features, normalization, and labels; MOMENT consumes the normalized tensor, whereas classical estimators consume its deterministic flattened view.
Thus, neither anomaly-ranking protocol shows a benefit from frozen MOMENT-small over models fitted to target-normal data.

\begin{table*}[t]
\centering
\caption{Selected anomaly-ranking and synthetic-diagnostic results. MIMII values are five-seed means. BDG2 values first average five injection seeds within each meter and then macro-average the 12 meters; they do not represent real-fault detection.}
\label{tab:representative-results}
\small
\setlength{\tabcolsep}{4.5pt}
\resizebox{\textwidth}{!}{%
\begin{tabular}{@{}l l l r r r r r@{}}
\toprule
Protocol & Model & Scope & AUROC & AUPRC & F1 & Precision & Recall \\
\midrule
C-MAPSS & TCN-AE & 100 engines OOF & 0.9570 & 0.8960 & 0.7959 & 0.7751 & 0.8239 \\
C-MAPSS & Isolation Forest & 100 engines OOF & 0.9521 & 0.8866 & 0.7847 & 0.7659 & 0.8129 \\
C-MAPSS & MOMENT Recon. & 100 engines OOF & 0.7310 & 0.3080 & 0.1190 & 0.2592 & 0.0790 \\
MIMII & OCSVM & 5 seeds, 320 files/seed & 0.6944 & 0.7277 & 0.6109 & 0.6334 & 0.5912 \\
MIMII & MOMENT Recon. & 5 seeds, 320 files/seed & 0.5501 & 0.5890 & 0.2232 & 0.7489 & 0.1325 \\
BDG2 diagnostic & LightGBM & 12 meters, 5 seeds & 0.7039 & 0.1639 & 0.1050 & 0.0653 & 0.3562 \\
BDG2 diagnostic & Ridge & 12 meters, 5 seeds & 0.6853 & 0.1549 & 0.1070 & 0.0738 & 0.3012 \\
BDG2 diagnostic & Chronos-T5 tiny & 12 meters, 5 seeds & 0.7215 & 0.1455 & 0.0699 & 0.0392 & 0.6358 \\
BDG2 diagnostic & TimesFM 2.5 & 12 meters, 5 seeds & 0.7239 & 0.1582 & 0.0732 & 0.0412 & 0.6293 \\
\bottomrule
\end{tabular}
}
\end{table*}

\subsection{Risk-Horizon Sensitivity}

The C-MAPSS ranking is stable across proxy horizons of 15, 20, 30, and 40 cycles.
TCN-AE AUROC ranges from 0.9347 to 0.9805 and AUPRC from 0.8725 to 0.8960; Isolation Forest remains close.
MOMENT reconstruction remains substantially lower at every horizon, with AUROC ranging from 0.7004 to 0.7682 and AUPRC from 0.1910 to 0.3646.
Although the proxy definition changes absolute performance, MOMENT remains behind the fitted references at every tested horizon.

\subsection{Forecasting and Perturbation Sensitivity}

On the aligned 12-meter BDG2 panel, TimesFM~2.5 has the lowest meter-macro forecast errors (MAE 9.3059, RMSE 15.7695, sMAPE 0.1033).
Chronos-T5 tiny reaches 10.3225/17.0758/0.2214, while Ridge and LightGBM have similar larger relative errors.
TimesFM and Chronos achieve aggregate synthetic AUROC of 0.7239 and 0.7215, compared with 0.7039 for LightGBM and 0.6853 for Ridge.
Aggregate AUPRC is close and ordered differently: 0.1582, 0.1455, 0.1639, and 0.1549, respectively.
This contrast illustrates why AUROC alone can overstate practical differences under sparse perturbations.

Perturbation-specific results delimit this evidence.
Missingness is perfectly separable by construction, while spike and drop events are comparatively easy.
Delayed response, flatline, stuck-at, and seasonal distortion have low AUPRC for all evaluated models.
Chronos has the highest meter-macro AUROC point estimates on gradual drift and level shift, closely followed by TimesFM, but thresholded precision remains low.
The aggregate diagnostic therefore mixes qualitatively different event difficulties and cannot establish real building-fault detection.

\subsection{Local Resource Requirements}

\begin{figure*}[t]
\centering
\includegraphics[width=0.88\textwidth]{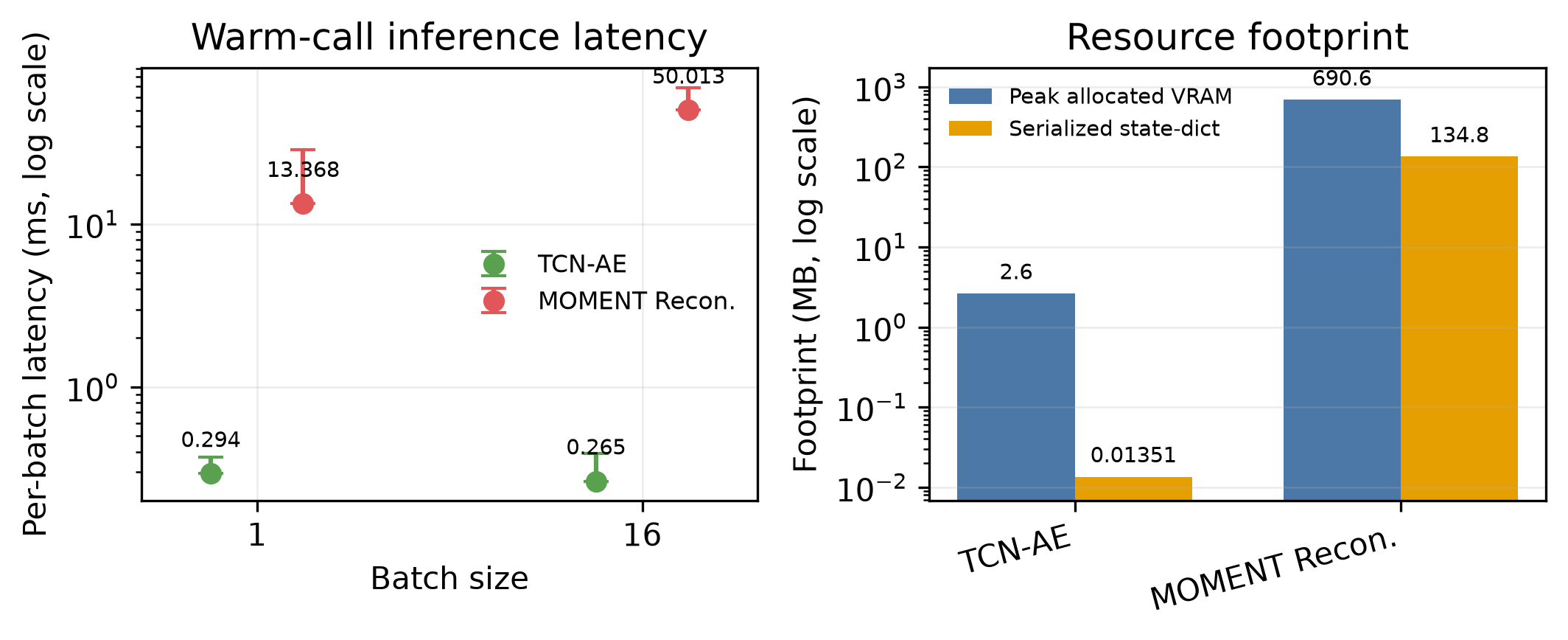}
\caption{Standardized local resource audit on one RTX 4090. The left panel reports same-process warm-call per-batch p50 latency and p50--p95 ranges over 30 repetitions after 10 warm-ups. The right panel reports peak allocated VRAM and serialized state-dictionary size. Preprocessing and model loading are excluded.}
\label{fig:cost-performance}
\end{figure*}

Figure~\ref{fig:cost-performance} compares TCN-AE and MOMENT reconstruction on the same RTX 4090 and scaled fold-0 workload.
At batch size 1, per-batch p50 latency is 0.294 ms for TCN-AE and 13.368 ms for MOMENT; at batch size 16 it is 0.265 and 50.013 ms.
MOMENT's serialized state-dictionary size is 138,053 KB versus 13.8 KB for TCN-AE, and its batch-16 peak allocated VRAM is 691 MB versus 2.6 MB.
These are same-process warm-call distributions on a shared host, which exhibited latency dispersion; they establish implementation-specific scale, not a stable speed multiplier or hardware-independent frontier.

\section{Discussion}
\label{sec:discussion}

\subsection{When Do TSFMs Pay Off?}

BDG2 is the only setting in which zero-shot forecasting priors improve selected point estimates relative to the fitted residual baselines, but model preference changes across AUROC, AUPRC, and thresholded metrics.
Because the labels are injected, this is evidence of perturbation sensitivity rather than real-fault detection.

The comparison evaluates deployment modes: TSFMs are frozen or zero-shot, whereas classical and compact neural models fit target-normal data.
It does not assess the full fine-tuning capacity of TSFMs.
Across the two anomaly-ranking protocols, avoiding TSFM adaptation does not offset lower ranking performance, and the audited MOMENT mode adds local implementation cost.
Model selection should therefore favor the lowest-cost model that satisfies prespecified protocol-specific performance requirements rather than treat pretraining as evidence of fitness by itself.

\subsection{Threats to Validity}

\paragraph{Task heterogeneity.}
C-MAPSS uses degradation-risk proxy labels, MIMII uses labeled acoustic anomalies, and BDG2 uses residual monitoring with synthetic sensitivity diagnostics.
The study is a protocol-aware audit, not a unified benchmark.

\paragraph{C-MAPSS label and split semantics.}
High-risk labels are derived from each engine's observed terminal cycle and are not native anomaly annotations.
Engine-disjoint folds prevent engine and exact-window leakage, but stride-1 windows remain highly dependent within an engine.
Within-fold engine-cluster bootstrap respects this grouping without comparing raw score ranks across fold-specific scales, and the four-horizon analysis shows that absolute performance remains proxy-dependent.

\paragraph{Synthetic energy anomalies.}
BDG2 pseudo metrics measure response to ten synthetic event types on 12 selected meters and five injection seeds.
Missingness is directly observable, while several contextual or temporal types remain difficult.
The selected panel broadens coverage beyond a single meter but is not a random sample of all buildings.
It does not measure real equipment faults, occupancy anomalies, commissioning problems, or meter failures.

\paragraph{TSFM coverage.}
We evaluate MOMENT-small, Chronos-T5 tiny, and TimesFM~2.5 200M.
Larger variants, fine-tuning, anomaly-specific adapters, and alternative context construction could change the results.
Our conclusion is conditional on the tested protocols.

\paragraph{Timing boundaries.}
The standardized C-MAPSS audit fixes workload, GPU, precision, batch size, warm-up, and repetition count, but excludes preprocessing and model loading.
The shared host shows nontrivial latency dispersion, and the CPU Isolation Forest configuration is not directly comparable to GPU implementations.

\paragraph{Uncertainty.}
C-MAPSS uses fold-stratified engine-cluster bootstrap and MIMII uses paired five-seed manifests with label-by-machine-ID-stratified file-bootstrap artifacts, but deep models are not repeatedly initialized within every fold.
BDG2 reports meter dispersion rather than a population-level confidence interval, and injection seeds are nested within the fixed panel.

\section{Conclusion}
\label{sec:conclusion}

Across engine-disjoint C-MAPSS and paired MIMII protocols, fitted lightweight models remain stronger than the tested frozen MOMENT-small modes, and this ordering persists across four C-MAPSS risk horizons.
Forecasting TSFMs show limited advantages on BDG2 in forecast error and synthetic AUROC, but not in AUPRC or thresholded F1; this is not evidence of real-fault detection.
Foundation models should therefore be evaluated as task-dependent deployment choices rather than default replacements for fitted models: compact fitted models are strong references when target-normal data and resource constraints matter, while zero-shot forecasters merit evaluation when residual forecasting is itself valuable.
Stronger claims require real energy-event labels and adapted or anomaly-specific TSFM evaluation.

In practice, deployment decisions should retain target-normal baselines and report calibration boundaries alongside ranking metrics.
Synthetic diagnostics can screen forecasters but cannot certify detectors without real event labels and end-to-end resource accounting.

\clearpage
\begingroup
\fontsize{9.5}{11.3}\selectfont
\setlength{\bibsep}{2.5pt plus 0.2ex}
\bibliographystyle{plainnat}
\bibliography{references}
\endgroup

\end{document}